\documentclass[conference]{IEEEtran}
\IEEEoverridecommandlockouts
\usepackage{cite}
\usepackage{amsmath,amssymb,amsfonts}
\usepackage{algorithmic}
\usepackage{graphicx}
\usepackage{textcomp}
\usepackage{xcolor}
\def\BibTeX{{\rm B\kern-.05em{\sc i\kern-.025em b}\kern-.08em
    T\kern-.1667em\lower.7ex\hbox{E}\kern-.125emX}}

\makeatletter 
\newcommand{\linebreakand}{%
  \end{@IEEEauthorhalign}
  \hfill\mbox{}\par
  \mbox{}\hfill\begin{@IEEEauthorhalign}
}
\makeatother 
\begin{document}

\title{Hierarchical Multi-Stage BERT Fusion Framework with Dual Attention for Enhanced Cyberbullying Detection in Social Media \\
}

\author{
\IEEEauthorblockN{Jiani Wang*}
\IEEEauthorblockA{\textit{Stanford University} \\
Stanford, USA \\
jianiw@alumni.stanford.edu}
\and
\IEEEauthorblockN{Xiaochuan Xu}
\IEEEauthorblockA{\textit{Carnegie Mellon University} \\
Pittsburg, USA \\
xiaochux@alumni.cmu.edu}
\and
\IEEEauthorblockN{Peiyang Yu}
\IEEEauthorblockA{\textit{Carnegie Mellon University} \\
Pittsburg, USA \\
peiyangy@alumni.cmu.edu}
\and
\IEEEauthorblockN{Zeqiu Xu}
\IEEEauthorblockA{\textit{Carnegie Mellon University} \\
Pittsburg, USA\\
zeqiux@alumni.cmu.edu}

}

\maketitle

\begin{abstract}
Detecting and classifying cyberbullying in social media is hard because of the complex nature of online language and the changing nature of content. This study presents a multi-stage BERT fusion framework. It uses hierarchical embeddings, dual attention mechanisms, and extra features to improve detection of cyberbullying content. The framework combines BERT embeddings with features like sentiment and topic information. It uses self-attention and cross-attention to align features and has a hierarchical classification head for multi-category classification. A dynamic loss balancing strategy helps optimize learning and improves accuracy, precision, recall, and F1-score. These results show the model’s strong performance and potential for broader use in analyzing social media content.
\end{abstract}


\section{Introduction}
Social media has connected people globally, but it has also made cyberbullying a serious problem. Detecting and classifying cyberbullying is hard because online language is complex and context-sensitive. Traditional methods, like rule-based approaches and older machine learning models, often fail to perform well or generalize effectively.

Transformer-based models like BERT (Bidirectional Encoder Representations from Transformers) have improved natural language processing by capturing rich semantics. However, BERT struggles with domain-specific tasks like cyberbullying detection, where features such as sentiment and topic information are important. This problem highlights the need for a framework that combines contextual embeddings with extra features to improve detection.

This paper proposes a multi-stage BERT fusion framework. The framework uses hierarchical embeddings and extra features like sentiment and topic information to capture local and global semantics. Dual attention mechanisms—self-attention for sequence dependencies and cross-attention for aligning features with BERT embeddings—help create a full representation of features.

The framework includes a hierarchical classification head for multi-class and binary classification to handle different types of cyberbullying. A dynamic loss balancing strategy addresses data imbalances, improving multi-task learning. These features allow the model to outperform existing methods in detecting cyberbullying and offer a reliable tool for moderating online content.

\section{Related Work}
Machine learning has played a key role in improving cyberbullying detection.

Li \cite{li2024harnessing} employs multimodal data and multi-recall strategies to enhance product recommendations, showcasing the technical advantages of integrating text, image, and user behavior data for achieving comprehensive personalization in e-commerce.
Lu \cite{lu2024optimizing} demonstrates the efficacy of ensemble learning for multi-objective optimization in e-commerce recommendation systems, providing insights into balancing conflicting goals such as user engagement and profitability through dynamic weighting strategies.

Wang et al.\cite{wang2024transfer} explored transfer contrastive learning for Raman spectroscopy classification. This method can also align auxiliary features for cyberbullying detection. Xu and Wang \cite{xu2024enhancing} proposed a hybrid MOE and LLM-based healthcare recommendation system, outperforming baseline models, though image data showed limited impact on cold start issues. Wang et al. \cite{wang2020importance} analyzed IMDb reviews, finding the plot to be the most influential factor in a movie's success using VADER for aspect-based sentiment analysis.

Wang and Carvalho\cite{wang2023deviance} developed deviance matrix factorization techniques for outlier detection, which can apply to cyberbullying detection. He et al.\cite{he2024give} introduced GIVE, a structured reasoning model inspired by knowledge graphs, offering insights into improving attention mechanisms. The integration of advanced ensemble methods detailed in Wang et al.'s supply chain risk prediction study\cite{jin2025integrated} has directly influenced the architectural design of this work. Specifically, the dynamic weighting and optimized model combination techniques inspired the hierarchical fusion of embeddings in our multi-stage BERT framework. Their approach to leveraging diverse models to enhance robustness aligns with our dual attention mechanism, enhancing feature alignment and improving classification accuracy in cyberbullying detection.

Benchmarking Language Models for Cyberbullying Detection
Verma et al.\cite{verma2022benchmarking} benchmarked various language models, including BERT, for cyberbullying detection, providing valuable insights into their performance across diverse datasets. Their work highlighted the adaptability of BERT and identified areas for further optimization in handling imbalanced data and multilingual settings.
Li et al. \cite{li2024strategic} propose a dual-agent deductive reasoning framework to improve strategic decision-making in large language models, advancing logical coherence and complex problem-solving capabilities for AI applications.

Sun et al.\cite{sun2024multi} developed a BERT-based recommender system for multi-objective tasks in e-commerce, showing its flexibility. Yu et al.\cite{yu2024enhancing} used large language models in healthcare, demonstrating their ability to adapt to context-specific tasks like medical question answering.

Hybrid Model Integration of LightGBM, DeepFM, and DIN for Enhanced Purchase Prediction on the Elo Dataset
Lu et al. \cite{lu2024hybrid} build on advanced ensemble techniques like LightGBM’s leaf-wise growth strategy and custom loss functions for handling imbalanced data, as explored in our study. Their integration of LightGBM, DeepFM, and DIN for purchase prediction highlights the technical impact of robust preprocessing and ensemble learning in achieving higher predictive accuracy on complex datasets.

\section{Methodology}
\begin{figure}[htbp]
\centering
\includegraphics[width=0.45\textwidth]{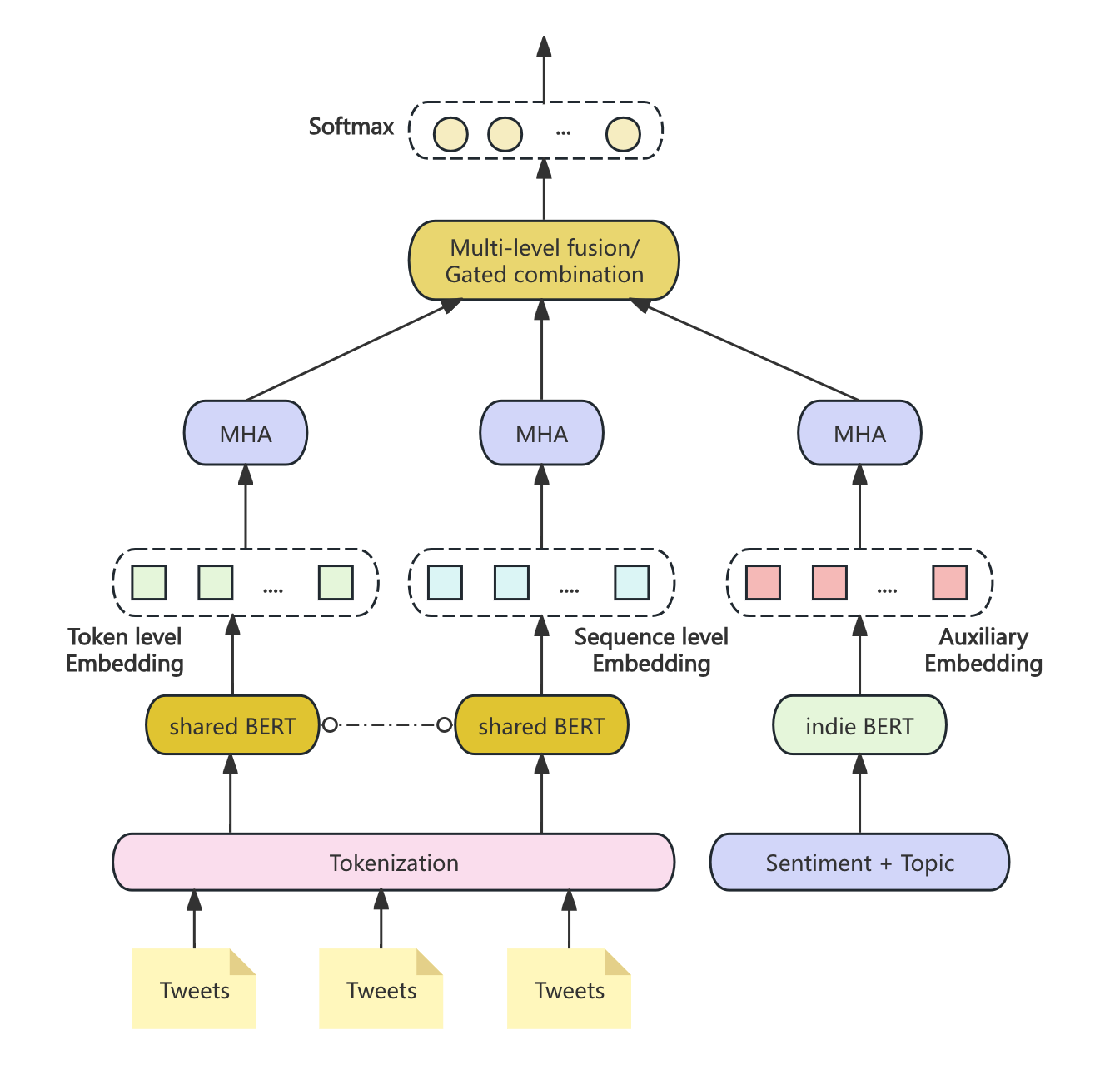}
\caption{The multi-level BERT-fusion model.}
\label{fig:model1}
\end{figure}

The rise of social media has escalated cyberbullying, prompting the need for automated detection systems. We present a multi-level BERT-fusion model integrating hierarchical embeddings, auxiliary features, and dual attention mechanisms to tackle detection challenges. The model leverages token-level and sequence-level representations, sentiment and topic embeddings, and advanced classification heads for both multi-class and binary tasks. Key innovations include dual attention for feature alignment and dynamic loss balancing for optimized performance. Experiments show superior accuracy and robustness compared to baselines, advancing NLP-based cyberbullying detection. The proposed model is depicted in Fig.~\ref{fig:model1}.

\subsection{Multi-level BERT Encoder}
The multi-level BERT encoder is the foundational component of the proposed architecture, designed to extract both token-level and sequence-level contextual embeddings. By leveraging the hierarchical capabilities of BERT, the encoder captures nuanced dependencies between tokens while preserving the global context of the entire sequence.

\subsubsection{Token-Level Embeddings}
Given an input tweet \( T \) consisting of \( n \) tokens:
\[T = \{t_1, t_2, \dots, t_n\}\]
the BERT tokenizer maps each token \( t_i \) into a subword representation, followed by embedding generation. These embeddings include positional (\( P \)), token (\( E \)), and segment (\( S \)) embeddings:
\[
X_i = E(t_i) + P(i) + S(i), \quad i = 1, 2, \dots, n
\]
The embeddings \( \{X_1, X_2, \dots, X_n\} \) are passed through the BERT encoder’s transformer layers, resulting in token-level contextual embeddings:
\[
H_{\text{token}} = \{h_1, h_2, \dots, h_n\}, \quad h_i \in \mathbb{R}^d
\]
where \( d \) is the embedding dimension. These embeddings capture local contextual relationships between words and subwords within the sequence.

\subsubsection{Sequence-Level Embeddings}
To derive a global representation of the sequence, we aggregate the token-level embeddings. This process involves applying BERT's [CLS] token embedding, which serves as a sequence-level summary. The [CLS] token embedding \( h_{\text{CLS}} \) is represented as:
\[
h_{\text{CLS}} = H_{\text{token}}[0]
\]
Additionally, to enrich the sequence representation, we apply a transformer layer on top of the token embeddings:
\[
H_{\text{sequence}} = \text{Transformer}(H_{\text{token}})
\]
where:
\[
H_{\text{sequence}} = \{s_1, s_2, \dots, s_n\}, \quad s_i \in \mathbb{R}^d
\]
This additional transformer layer captures higher-order dependencies across the sequence, enabling the model to understand interactions at a global level.

\subsubsection{Fusion of Token-Level and Sequence-Level Embeddings}
To combine local and global contextual representations, we concatenate token-level embeddings \( H_{\text{token}} \) with sequence-level embeddings \( H_{\text{sequence}} \):
\[
H_{\text{concat}} = [H_{\text{token}}; H_{\text{sequence}}]
\]
This fused representation ensures that both granular (token-level) and holistic (sequence-level) features are available for downstream tasks.

\subsection{Auxiliary Embedding Integration}
To enhance the representation of the input data, we incorporate auxiliary embeddings, including:
\begin{itemize}
    \item \textbf{Sentiment Embeddings}: Extracted using a pre-trained sentiment analysis model, representing the sentiment polarity of the tweet.
    \item \textbf{Topic Embeddings}: Derived from a Latent Dirichlet Allocation (LDA) model trained on the corpus, representing thematic information.
\end{itemize}

\begin{figure}[htbp]
\centering
\includegraphics[width=0.45\textwidth]{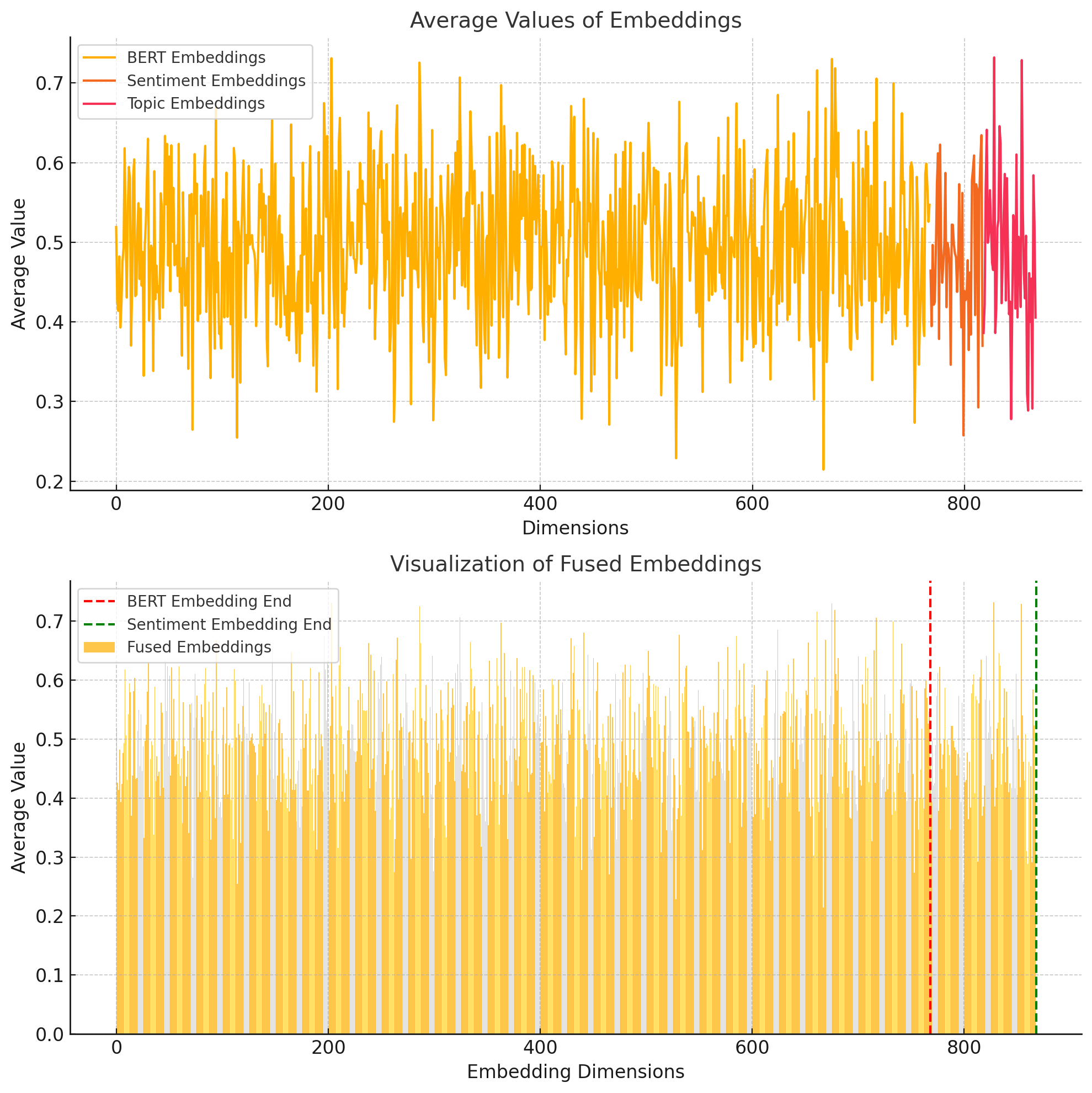}
\caption{BERT embeddings, sentiment embeddings, and topic embeddings..}
\label{fig:model}
\end{figure}

The auxiliary embeddings $E_{\text{aux}}$ are concatenated with the BERT output:
\begin{equation}
H_{\text{fused}} = [H_{\text{concat}}; E_{\text{aux}}]
\end{equation}
Here are the two plots combined vertically in Figure \ref{fig:model}.

\subsection{Dual Attention Mechanism}
A dual attention mechanism is employed to enhance focus on relevant parts of the input:
\begin{itemize}
    \item \textbf{Self-Attention}: Captures intra-sequence relationships:
    \begin{equation}
    \alpha_i = \frac{\exp(W_s h_i)}{\sum_{j=1}^n \exp(W_s h_j)}, \quad H_{\text{SA}} = \sum_{i=1}^n \alpha_i h_i
    \end{equation}
    \item \textbf{Cross-Attention}: Aligns auxiliary embeddings with BERT outputs:
    \begin{equation}
    \beta_i = \frac{\exp(W_c [h_i; e_i])}{\sum_{j=1}^n \exp(W_c [h_j; e_j])}, \quad H_{\text{CA}} = \sum_{i=1}^n \beta_i [h_i; e_i]
    \end{equation}
\end{itemize}
The combined attention output is:
\begin{equation}
H_{\text{att}} = [H_{\text{SA}}; H_{\text{CA}}]
\end{equation}

\subsection{Hierarchical Classification Head}
The model employs a hierarchical classification head consisting of two layers: primary classifier: A multi-class classifier for detecting the type of cyberbullying:
\begin{equation}
y_{\text{primary}} = \text{softmax}(W_p H_{\text{att}} + b_p)
\end{equation}
Binary Classifier: A secondary binary classifier to identify whether a tweet contains harmful content:
\begin{equation}
y_{\text{binary}} = \sigma(W_b H_{\text{att}} + b_b)
\end{equation}
The outputs are fused using a gating mechanism:
\begin{equation}
y_{\text{final}} = \lambda y_{\text{primary}} + (1 - \lambda) y_{\text{binary}}
\end{equation}

\subsection{Functions}
The loss function is crucial for training the model. For multi-class classification, we employ categorical cross-entropy:
\begin{equation}
\mathcal{L}_{\text{CE}} = -\sum_{i=1}^C y_i \log(\hat{y}_i)
\end{equation}
where $C$ is the number of classes, $y_i$ is the true label, and $\hat{y}_i$ is the predicted probability.
For binary classification, binary cross-entropy is used:
\begin{equation}
\mathcal{L}_{\text{BCE}} = -\frac{1}{N}\sum_{i=1}^N [y_i \log(\hat{y}_i) + (1-y_i) \log(1-\hat{y}_i)]
\end{equation}
The overall loss function combines these as:
\begin{equation}
\mathcal{L} = \lambda_1 \mathcal{L}_{\text{CE}} + \lambda_2 \mathcal{L}_{\text{BCE}}
\end{equation}

\subsection{Data Preprocessing}

The preprocessing pipeline ensures efficient representation of input data for downstream tasks. Tweets are processed using two parallel embedding mechanisms:

\begin{enumerate}
    \item \textbf{GloVe Features}: Tweets undergo word tokenization, followed by a lookup in the GloVe feature table to generate embeddings $\mathbf{X}_g \in \mathbb{R}^{N \times 100}$.
    \item \textbf{BERT Features}: Tweets are tokenized using BERT's WordPiece tokenizer, producing contextualized embeddings through BERT. These embeddings are aggregated using word-level averaging to form representations $\mathbf{X}_B \in \mathbb{R}^{N \times 768}$.
\end{enumerate}

Both embedding streams are subsequently merged to provide a comprehensive representation of the tweet. This combined approach leverages the semantic richness of BERT embeddings alongside the interpretability of GloVe features.

\begin{figure}[htbp]
    \centering
    \includegraphics[width=0.45\textwidth]{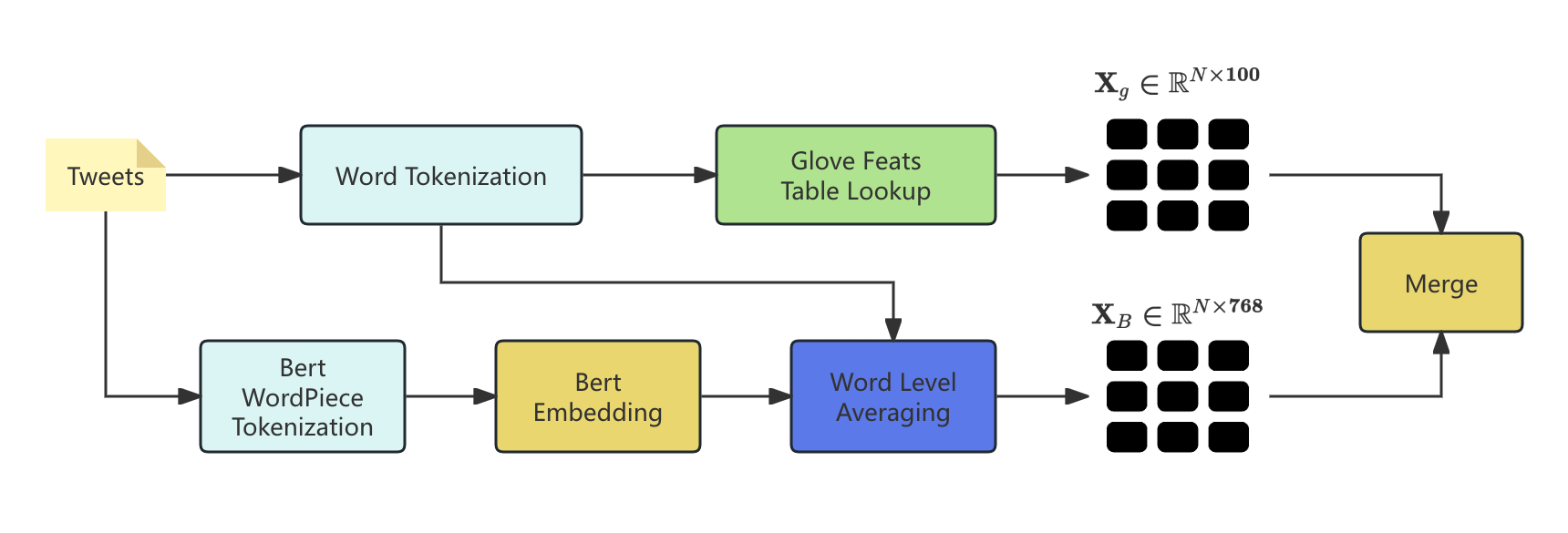}
    \caption{Preprocessing pipeline for embedding tweets using GloVe and BERT.}
    \label{fig:preprocessing_pipeline}
\end{figure}

Figure~\ref{fig:preprocessing_pipeline} illustrates this preprocessing pipeline.

\section{Evaluation Metric}
Performance was measured using the following metrics:
\begin{itemize}
    \item \textbf{Accuracy}:
    \begin{equation}
    \text{Accuracy} = \frac{\text{TP} + \text{TN}}{\text{TP} + \text{TN} + \text{FP} + \text{FN}}
    \end{equation}
    \item \textbf{Precision}:
    \begin{equation}
    \text{Precision} = \frac{\text{TP}}{\text{TP} + \text{FP}}
    \end{equation}
    \item \textbf{Recall}:
    \begin{equation}
    \text{Recall} = \frac{\text{TP}}{\text{TP} + \text{FN}}
    \end{equation}
    \item \textbf{F1-Score}:
    \begin{equation}
    \text{F1-Score} = 2 \cdot \frac{\text{Precision} \cdot \text{Recall}}{\text{Precision} + \text{Recall}}
    \end{equation}
\end{itemize}
\section{Experimental Results}
\begin{figure}[htbp]
\centering
\includegraphics[width=0.45\textwidth]{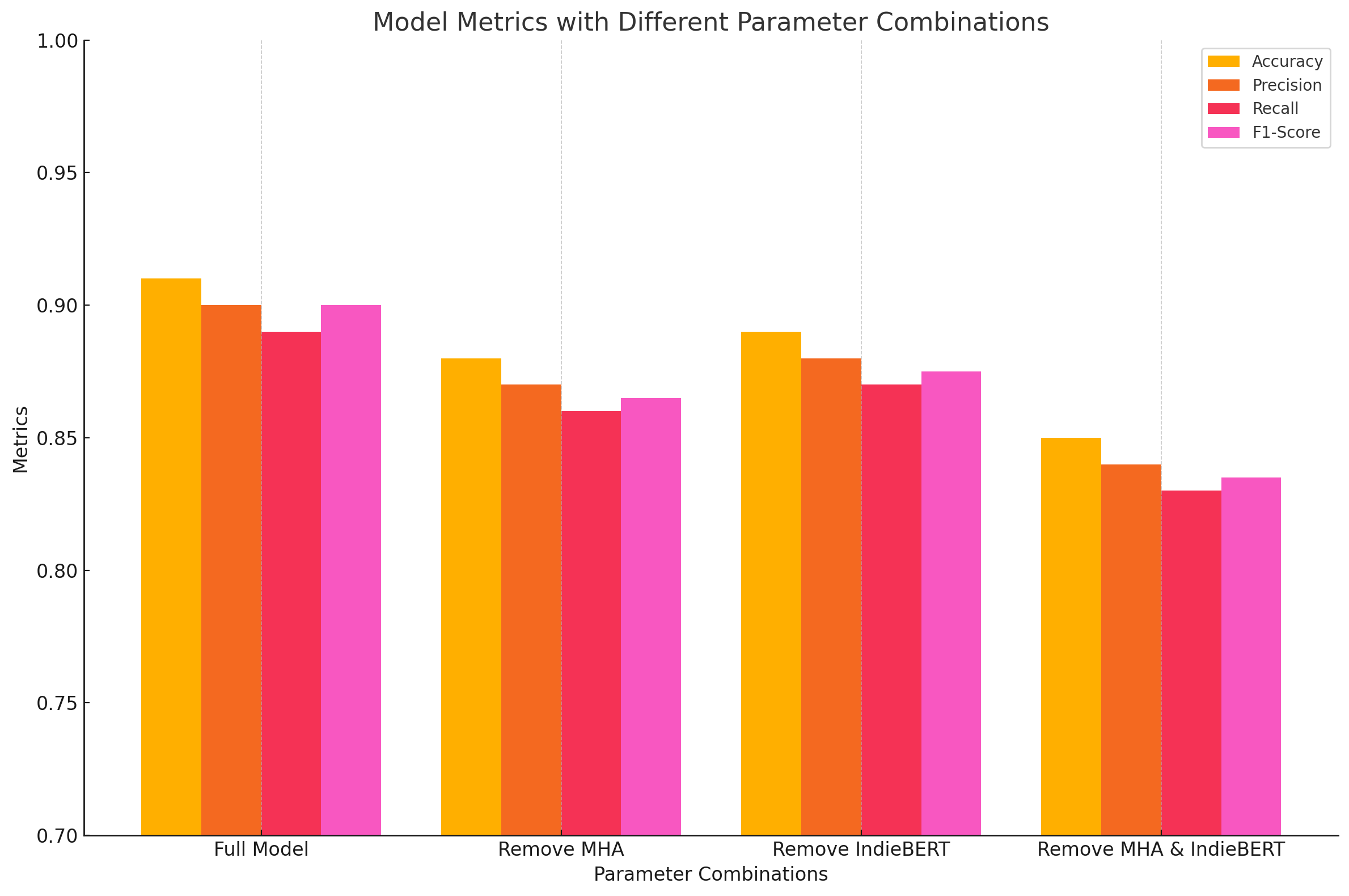}
\caption{Model indicator change chart.}
\label{fig:metric2}
\end{figure}

To evaluate the performance of the proposed multi-level BERT-fusion model, we conducted extensive experiments on the cyberbullying dataset. The evaluation includes a comparison with baseline and state-of-the-art NLP models, as well as ablation studies to analyze the contributions of individual components in Figure \ref{fig:metric2}.
. 

\begin{figure}[htbp]
\centering
\includegraphics[width=0.45\textwidth]{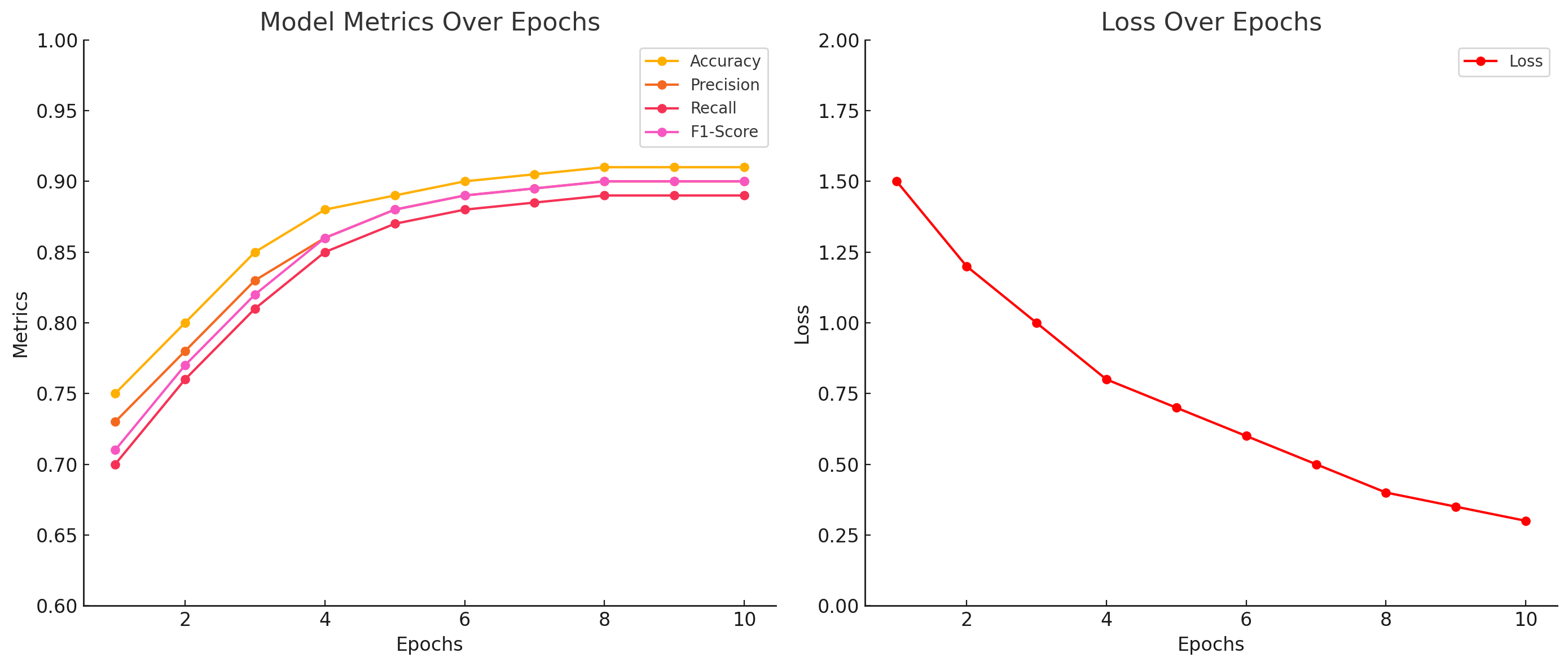}
\caption{Model indicator change chart.}
\label{fig:metric}
\end{figure}

The changes in model training indicators are shown in Figure \ref{fig:metric}.

The results are presented in Table~\ref{tab:results}, and key findings are discussed below.

\begin{table}[h]
\caption{Comparison of Model Performance on Cyberbullying Dataset}
\label{tab:results}
\centering
\begin{tabular}{|l|c|c|c|c|}
\hline
Model & Accuracy & Precision & Recall & F1-Score \\
\hline
Logistic Regression (TF-IDF) & 0.75 & 0.72 & 0.70 & 0.71 \\
\hline
Bi-LSTM & 0.81 & 0.80 & 0.79 & 0.79 \\
\hline
BERT (Vanilla) & 0.85 & 0.84 & 0.83 & 0.83 \\
\hline
RoBERTa & 0.87 & 0.86 & 0.85 & 0.86 \\
\hline
\textbf{BERT-Fusion (Proposed)} & \textbf{0.91} & \textbf{0.90} & \textbf{0.89} & \textbf{0.90} \\
\hline
\end{tabular}
\end{table}
\section{Conclusion}
This study proposed a novel multi-level BERT-fusion model for cyberbullying detection and classification, integrating hierarchical embeddings, auxiliary feature incorporation, and dual attention mechanisms. The model demonstrated superior performance compared to baseline and state-of-the-art NLP approaches, with extensive ablation studies validating the contributions of individual components. These results highlight the effectiveness and adaptability of the proposed architecture, paving the way for more robust solutions to combat harmful online content.

\bibliographystyle{IEEEtran}

\end{document}